# A Classification Supervised Auto-Encoder Based on Predefined Evenly-Distributed Class Centroids

Qiuyu Zhu [a], Ruixin Zhang [a]

[a] School of Communication and Information Engineering, Shanghai University, Shanghai 200444, China

**Abstract** Classic variational autoencoders are used to learn complex data distributions, that are built on standard function approximators. Especially, VAE has shown promise on a lot of complex task. In this paper, a new autoencoder model - classification supervised autoencoder (CSAE) based on predefined evenly-distributed class centroids (PEDCC) is proposed. Our method uses PEDCC of latent variables to train the network to ensure the maximization of inter-class distance and the minimization of inner-class distance. Instead of learning mean/variance of latent variables distribution and taking reparameterization of VAE, latent variables of CSAE are directly used to classify and as input of decoder. In addition, a new loss function is proposed to combine the loss function of classification. Based on the basic structure of the universal autoencoder, we realized the comprehensive optimal results of encoding, decoding, classification, and good model generalization performance at the same time. Theoretical advantages are reflected in experimental results.

## 1. Introduction

Over the past few years, variational autoencoder have demonstrated their effectiveness in many areas, such as unsupervised learning [1], supervised learning [2], semi-supervised learning [3]. The theory of variational autoencoder is from the perspective of Bayesian Theorem, the posterior distribution of the latent variables **z** conditioned on the data **x**, is approximated by a normal distribution, who's mean and variance are the output of a neural network. To make the generated sample **x*** very similar to data **x**, VAE adds Kullback-Leibler divergence to the loss function, where the latent variables **z** mapped from the data **x** corresponds to encoder, and the sample **x*** generated from the distribution of latent variables **z** corresponds to decoder. The complex mapping function is learned by neural network, as the neural network can approximate all functions theoretically.

Taking the classification as an example, we want to use high dimensional hidden variables to represent the input data. In the high-dimensional space, the ultimate objective of classification is making inter-class distance to be as large as possible, that is, the sample is more separable, and the inner-distance within the same class is as small as possible. In other words, the distribution of each class is more aggregated. For the distribution of the original data, in the case of MNIST [11], the distribution of the number "9" is closer to the number "4" and the number "7" [2, 22], that's the reason why CVAE mix them sometimes.

A lot of work has been done to prove the effectiveness of codec structure. Based on the basic structure of autoencoder and predefined evenly-distributed class centroids (PEDCC), a new supervised learning method for autoencoder is proposed in this paper. By PEDCC that meets the criteria of maximized inter-class distance (the distance between classes is the furthest), we map the label of input data **x** to different predefined class centroids, then let the encoder network learn the mapping function. Through the joint training, the network mapping makes the latent features of the same class samples as close as possible to predefined class centroids, finally to get a good classification. As far as we know, prior to this article, there was no method of using predefined class centroids to train automatic encoders.

We use the output of the encoder as input to the decoder directly, and to make the autoencoder input and output as close as possible, where the mean square error (MSE) loss function is adopted. Because of resampling, the image quality generated by VAE generally has the problem of edge blurring. To solve this problem, we added the wavelets loss, that is, a wavelets transform is taken on the input image and the output of the autoencoder respectively, whose difference is taken as a new loss function to improve the edge quality of the generated image. This is an additional constraint to the edge difference of the input and generated image, which is more conducive to improving the subjective quality of the image. To further improve the subjective quality of the generated image, we draw on the idea of reparameterization in VAE, to add Gaussian noise to the latent features to reconstruct the input of the decoder in training phase. The experiment results prove that this trick is very effective.

Our main contributions are as follows:

1) The PEDCC is proposed to meet the criteria of maximized inter-class distance, so that convolutional neural networks can focus more on learning more compact intra-class distances.

2) By PEDCC, our method combines classification and autoencoder, latent variables can be used both for classification and reconstruction, and add noise during training to improve the accuracy of classification and image quality simultaneously.

3) To further improve image quality, wavelets loss function is proposed. By combining traditional pattern recognition methods, a constraint is placed on both the high-frequency and low-frequency information of the image, which also good for classification and reconstruction performance.

Below, we first introduce some of the previous work in Section 2. In section 3 our approach is described in detail. Then in section 4 we will verify the validity of our method through the experimental results on different datasets. Finally, in section 5, we discuss some of the issues that still exist and what will be done in the future.

## 2. Related work

The autoencoder is an unsupervised learning algorithm, which is mainly used for dimension reduction or feature extraction. It also can be used in deep learning to initialize weight before training phase. Depending on different application scenarios, autoencoders can be divided into sparse autoencoder [13,14], which is add L1 regularization to the basic autoencoder to make feature sparse, denoising autoencoder [15,16], which is designed to prevent the overfitting problem and add noise to the input data, to enhance the generalization ability of the model and variational autoencoder [1,18], which learn the distribution of raw data by setting the distribution of latent variables as $\mathcal{N}(0,I)$, and can in turn produce some data similar to the original data. Normally, the variational autoencoder is used in unsupervised learning, so we cannot control the generation of the decoder. What's more, the conditional variational autoencoder (CVAE) [19,20] combines the variational autoencoder with supervised information, which allows us to control the generation of decoder. Ref. [2] assumes that class labels are independent of latent features so that they are stitched together directly to generate data. Ref. [24] controls the generation of latent variables through the labels of the face properties, and then generates data by sampling directly from the distribution of latent variables. Ref. [19] considers the prediction problem directly, with the label **y** as the data to be generated, and data **x** as a label, to achieve the purpose of predicting the label y of data **x**. [28,29] train generator directly, the latent variables **z** is also trained as the network parameters. Therefore, they only can generate images, without the function of feature extraction and classification.

The above works only learn the data distribution and complete codec work, without classification function. In recent years, some scholars use VAE in the field of incremental learning, which is mainly focus on how to alleviate catastrophic forgetting [25]. Ref. [21] adds a classification layer to the VAE structure, and use dual network structure, composed of "teacher-student" model to mitigate the problem of catastrophic forgetting. To achieve the classification function, based on CVAE, Ref. [22] adds an additional classification network for joint training. Usually, to add a classification function to an autoencoder, an additional network structure is necessary.

Through predefined evenly-distributed class centroids, CSAE proposed in this paper maps training labels to these class centers and can use latent variables to classify directly. In view of this new framework, a new loss function is also proposed for training.

## 3. Method

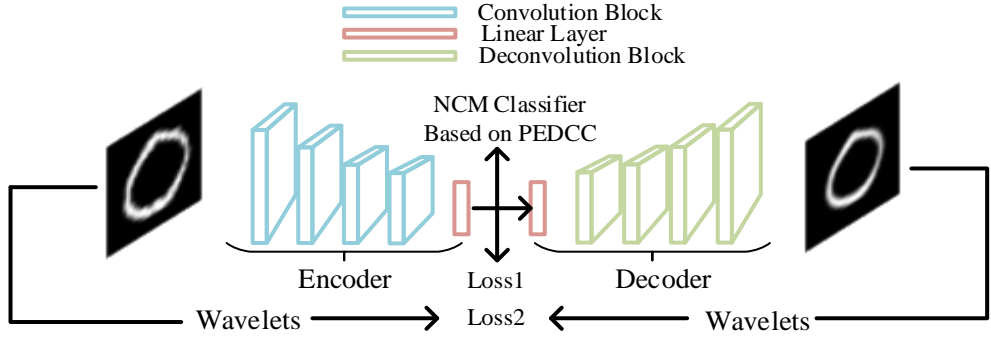

**Fig.1.** The structure of CSAE

In this section we will introduce the details of CSAE and show how to combine them to form an end-to-end learning system. Section 3.1 gives a description of the PEDCC. Section 3.2 describes the loss function, and section 3.3 discusses the network structure.

### 3.1 Predefined Evenly-Distributed Class Centroids

From the traditional view of statistical pattern recognition, the main objective of dimension reduction is to generate low-dimensional expressions with maximum inter-class distance and minimum inner-class distance, such as LDA algorithm. For deep learning classification model, the last Softmax is a linear classifier, while the preceding multilayer network is a mapping of dimension reduction, to generate low-dimensional latent variables. If the generated sample latent variables have the characteristics of small inner-class and large inter-class distance, neural networks will learn better features. Ref. [4] and [5], modify the Softmax function of cross-entropy loss in convolution neural networks (CNNs) to improve classification performance. Because of strong learning abilities, it is not difficult for neural networks to obtain a good aggregation within the same class. However, maximizing the inter-class distance is a difficult problem, it varies in different classification tasks. If the variance within the class is large and the inter-class distance is close, there will be overlaps between different classes, which can lead to wrong classification. There is no good way to avoid this problem. In this work, the class center of latent variables is artificially set by the method of PEDCC to make sure the distance of these clustering centers is furthest.

Thus, we learn a mapping function through the encoder in the autoencoder, and map the different classes of samples to these predefined class centers respectively, so that the distance between different classes can be separated by the strong fitting ability of deep learning, and the validity of the PEDCC is proved by experiments in this paper.

First, we assume that all clustering centers are distributed on a hypersphere, with the goal of generating $n$ points that evenly distributed on the d-dimensional hypersphere surface as clustering centers, where *n* is the number of classes. There is no analytical solution to generate evenly-distributed points on the hyperspherical surface, and even if we can generate, there are infinite solutions. Generally, the numerical solutions are obtained by iterative method. For us, as long as these center points are evenly distributed, then deep learning wound have sufficient fitting capability to achieve this mapping.

**Table 1** PEDCC algorithm

| Algorithm1 PEDCC |
| --- |
| **input** number of classes *n* |
| **input** Number of Iterations q |
| **require** class centroids $\boldsymbol{u_i} \in N(0, I)$, i = 1, ... , *n* |
|     $(\boldsymbol{u_1}, ..., \boldsymbol{u_n})$ ← Random generate class centroids |
|     U= $(\frac{u_1}{\|u_1\|}, ..., \frac{u_n}{\|u_n\|})$ ← Normalized $u_i$ |
|     V= $(\boldsymbol{v_1}, ..., \boldsymbol{v_n})$ ← Initialization velocity Parameter |
|     **for** k = 1,..., q **do** |
|         L ← Compute the distance between $u_i$ |
|         $F_v$ ← Compute resultant force tangent component |
|         U ← Update class centroids |
|         V ← Update velocity Parameter |
|     **end for** |
| **return** U= $(\boldsymbol{u_1}, ..., \boldsymbol{u_n})$ |

In this paper, the method of PEDCC is based on the physical model with the lowest like charge energy on the hyperspherical surface. That is, the *n* charge points on the hyperspherical surface have the repulsive force between each other, and the repulsive force push the charges to move. When the motion is finally balanced and the points on the hypersphere surface stop moving, *n* points will eventually be evenly distributed on the hyperspherical surface. When the equilibrium state is reached, the *n* points can be the furthest apart. We give each point a variable *v* to describe the state of motion, each step of the iteration updates the motion state and position of all points.

If the size of the input images is *m*×*m*, the dimension of the encoder output feature of the autoencoder is *d*, and the number of classes is *n*. To ensure that the distance between the predefined center points in the iteration is not too close, the PEDCC algorithm needs to set the distance threshold *ε*. When the distance between two points is less than *ε*, it is set to ε, then the iteration continues. Here, we set *ε* to 0.01.

The detailed algorithm is shown in Table 1, The algorithm requires the output of *n* points of d-dimensions predefined class centroids $(\boldsymbol{u_1}, ..., \boldsymbol{u_n})$. First, it is necessary to randomly sample *n* vectors in the *d*-dimensional normal distribution to represent the initial value of the predefined center, then these vectors are normalized to 1. Each point has a speed state $(\boldsymbol{v_1}, ..., \boldsymbol{v_n})$, to describe their state of motion. Then we go into the iterations: the distance matrix, between the initial n vectors are calculated firstly, and then calculate the resultant force $\boldsymbol{f}$ of each class center by others points according to the distance matrix. To simplify the calculation process, we assume $f \propto \frac{1}{l_{ij}^2}$, where $\boldsymbol{l_{ij}}$ is the distance vector between two points *i*, *j*. The resultant force vector for *i*-th points is:

$$\boldsymbol{f_i} = \sum_{j=0}^{n} \frac{1}{|l_{ij}|^2} \frac{l_{ij}}{|l_{ij}|} \quad j \neq i \tag{1}$$

With the resultant force, we can get the tangent vector of the resultant force $\boldsymbol{f_t}$:

$$\boldsymbol{f_t} = \boldsymbol{f} - dot(\boldsymbol{u}, \boldsymbol{f})\boldsymbol{u} \tag{2}$$

where $dot(\cdot)$ means dot product between two vectors. Finally, the position of each point $\boldsymbol{u}_i$ is updated by the current velocity vector $\boldsymbol{v}_i$, and the speed according to the tangent vector of the resultant force $\boldsymbol{f}_t$ is also updated. where means dot product between two vectors. Finally, the position of each point $\boldsymbol{u}_i$ is updated by the current velocity vector $\boldsymbol{v}_i$, and the speed according to the tangent vector of the resultant force $\boldsymbol{f}_t$ is also updated.

$$\boldsymbol{u}^{k+1} = \boldsymbol{u}^k + \boldsymbol{v}^k \tag{3}$$

$$\boldsymbol{v}^{k+1} = \boldsymbol{v}^k + \lambda \boldsymbol{f}_t^k \tag{4}$$

where $k$ is iteration times, $\lambda$ is a constant, just like learning rate, here $\lambda=0.01$. Subsequently, $\boldsymbol{u}$ is normalized. The algorithm ends until the maximum number of iterations is reached.

To verify the validity of the algorithm 1, we set the dimension of the latent variables to three dimensions for easy display. Figure 2 shows the distribution of 2, 4, 10 and 20 points respectively. The distribution of predefined centers on the spherical surface can be seen, the results are very close to evenly distributed. The result of our algorithm 1 is the optimal solution, which lays an important foundation for the subsequent classification and the generation of random sample.

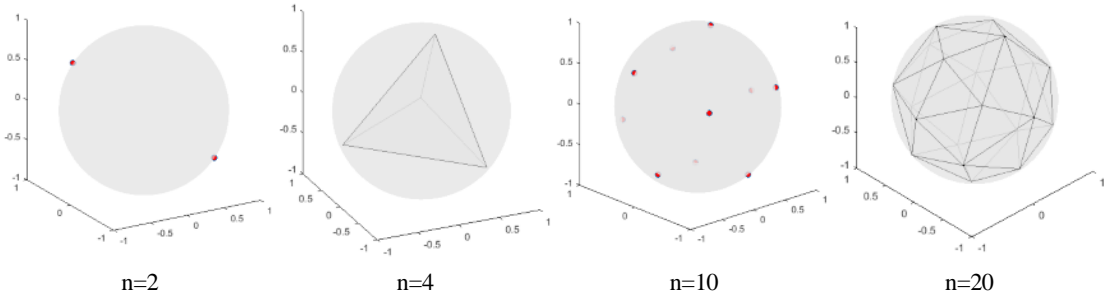

**Fig.2.** The distribution of points on sphere

The initial values of predefined centers here are randomly generated, and the convergence results are affected by the initial value, but the final distribution will be distributed evenly. Once determined, it is no longer changed in the later network training and test phase. Since the vector values of points on the high-dimensional sphere are too small for each dimension, to facilitate the subsequent works, we multiply them with a constant coefficient alpha to enlarge the range of values, here we take:

$$\alpha = \sqrt{d} \tag{5}$$

### 3.2 Design of Loss Function

#### 3.2.1 Loss function of classification

To use cross-entropy function as the last classification loss function, the outputs of neural network for classification [6,7,8] are usually mapped to one-hot type. The biggest difference of our framework is that the goal of optimization is that each class centroid is as close as possible to the corresponding predefined class centroid, which is a multidimensional vector, and the dimensions here are consistent with the feature's dimension of the encoder output, i.e. latent variables. The number of latent variables is normally greater than the number of classes, so that the cross-entropy loss function can no longer be used. In this paper, the mean square error function is used as the optimization target of our classification task.

Assume that the mapping function learned by the encoder is:

$$z = f_{encoder}(\boldsymbol{X}), \ \boldsymbol{X} \epsilon \mathbb{R}^{m \times m}, z \in \mathbb{R}^d$$

Then, the loss function of classification is:

$$Loss1(\mathbf{z}, \boldsymbol{\mu}) = \frac{1}{n}\sum_{i=1}^{n}(z_i - \mu_i)^2 \qquad (6)$$

where $\boldsymbol{\mu}_i$ is the predefined centroids of class *i*.

**3.2.2 Loss function of classification**

Supposing that the mapping function of the learned decoder is:

$$\mathbf{X}^* = f_{decoder}(\mathbf{z}), \mathbf{X}^* \in \mathbb{R}^{m \times m}$$

Normally, to make the decoder's output **x\*** as similar as the input image **x**, the loss function of image codec calculates the mean square error between the input image and the output image pixel-wisely.

$$MSE(\mathbf{X}^*, \mathbf{X}) = \frac{1}{m^2}\sum_{i=1}^{m}\sum_{j=1}^{m}(X_{ij}^* - X_{ij})^2 \qquad (7)$$

where m is the image dimension.

Because of the resampling of VAE generation model and the use of mean square error as the basis of image similarity measurement, the generated images of VAE model have obvious blurring phenomena. To improve the image quality, the operation of Wavelets Transform is added in this paper. Ideally, the generated image should be consistent with the input image, and the consistency here should also include clear edges.

To generate better quality images, Laplacian pyramid loss had been applied in [29]. Because we add noise to latent variables in the training phase, and Laplacian operator is very sensitive to noise, it is no longer applicable here. We choose Wavelets transform, which carries out low-pass and high pass filtering from both horizontal and vertical directions. As shown in Fig.3. b1 represents the image down sampled 2×, h1, v1, c1 represent the details in horizontal, vertical, and diagonal direction, respectively. After wavelets transform of input image and output image, we directly make MSE for the down sampled image and L1 regularization for the horizontal, vertical and diagonal image details respectively. The gradient of L2 regularization is related to the loss value, with the decrease of the value of loss, the gradient will be smaller and smaller, while the gradient of L1 regularization will always be 1, which is more beneficial for us to get sharper edges.

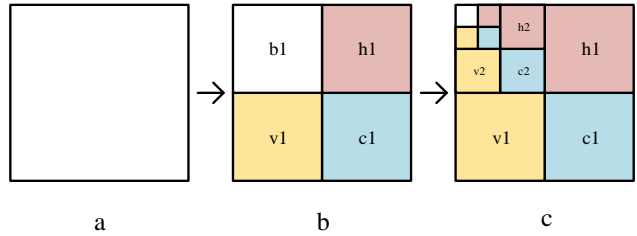

a    b    c

**Fig.3.** Wavelets transform

$$b_J, h_q, v_q, c_q = W^J(\mathbf{X}) \quad q \in 1,2,\dots,J$$

$$Loss2(\mathbf{X}^*, \mathbf{X}) = \lambda_1 MSE(b_J^*, b_J) + \lambda_2 \sum_{q=1}^{J} L1(h_q^*, h_q) + \lambda_3 \sum_{q=1}^{J} L1(v_q^*, v_q) + \lambda_4 \sum_{q=1}^{J} L1(c_q^*, c_q) \qquad (8)$$

where $\lambda_1, \lambda_2, \lambda_3, \lambda_4$ are constantly used to balance the relationship between the three loss functions. We set $\lambda_1 = 1, \lambda_2 = 1, \lambda_3 = 6, \lambda_4 = 10$ in training phase. According to the different resolutions of the picture, $J = 2$ on the MNIST / FashionMNIST dataset and $J = 3$ on the FaceScrub dataset.

$$L1(\mathbf{X}^*, \mathbf{X}) = \frac{1}{m^2}\sum_{i=1}^{m}\sum_{j=1}^{m}|X_{ij}^* - X_{ij}| \qquad (9)$$

So, the final loss function is:

$$Loss = Loss1 + loss2 \qquad (10)$$

The loss function of CSAE no longer contains Kullback-Leibler divergence items, which reflects that this article attempts to train autoencoder from a new perspective. That is, unlike the VAE model based on Bayesian Theorem, the input and output of encoder and decoder here are just a complex nonlinear mapping to be learned, and is independent each other statistically.

**3.3 Architecture**

In recent years, with the rapid development of convolution neural networks, many excellent networks have been proposed, such as [9,10], and a great deal of works has proved their superiority. To extract more effective features, unlike [2], this paper uses the shortcut connection convolution network structure proposed by [8] to replace the conventional fully connected layer to build autoencoder, that is, the convolution block + linear layer to construct the encoder, and the linear layer + deconvolution block (linear interpolation + convolutional layer) to form the decoder. The specific structure is shown in Figure 1, and the output of the encoder is the input of the decoder. Here, the structure of each convolution block is residual bloc. CSAE is not limited to this structure.

Because the MNIST dataset is rather simple, when we reduce the input image to the $4 \times 4$ resolution, only one full-connection layer is used to map the feature directly to the $d$ dimension latent variables. The decoder is also a fully connected layer + some deconvolution blocks, where the last layer of the decoder is a $1 \times 1$ convolution layer to change the number of output channels directly, and without the Sigmoid layer, which is also a different point between this paper and the CVAE structure. Table 2 is the network structure of CSAE used in this paper.

**Table 2** Structural details of CSAE

| Network | Operation | Input dims | Output dims |
|---|---|---|---|
| Encoder | Convolution | 3×32×32 | 32×32×32 |
| | Residual block | 32×32×32 | 64×16×16 |
| | Residual block | 64×16×16 | 128×8×8 |
| | Residual block | 128×8×8 | 128×4×4 |
| | Fully-connected | 128×4×4 | 40 |
| Decoder | Fully-connected | 40 | 128×4×4 |
| | Deconvolution block | 128×4×4 | 128×8×8 |
| | Deconvolution block | 128×8×8 | 64×16×16 |
| | Deconvolution block | 64×16×16 | 32×32×32 |
| | Convolution | 32×32×32 | 3×32×32 |

**3.4 Adding Noise to Latent Variables in Training Phase**

In VAE, the input of the decoder is a sampling of the distribution of the latent variables, this process is not a continuous operation and there is no gradient, during the training of stochastic gradient descent algorithm, backward propagation cannot continue to propagate. To solve this problem, VAE put forward the reparameterization trick, and a sampling is taken from normal distribution $\mathcal{N}(0, I)$. The sampled points are multiplied by the standard deviation, plus the mean value.

CSAE use the output of the encoder directly as the input of the decoder, and no longer requires reparameterization as [1][2]. However, we are inspired by reparameterization processing, where mean value can be considered to be the output of encoder, and the standard deviation is equivalent to a noise we add. The addition of noise in the training phase can make the decoder more insensitive to the change of input features, whose generalization ability (including interpolation and extrapolation) performs better and generates a more stable decode result. Based on the above understanding, before inputting the latent features inputs to the decoder, CSAE randomly generates a $d$-dimensional noise $n_0$ from $\mathcal{N}(0,I)$, and add it to the latent variable as the input to the decoder. As mentioned in section 3.1, after normalized the predefined center to the surface of hypersphere, we also multiply them all by a constant coefficient $\alpha$ to expand the latent variable value space.

Similarly, to adapt to the increase of the latent variable value space, we also amplify $\boldsymbol{n_0}$ which meet the standard normal distribution, the amplification of the noise is:

$$\boldsymbol{n_0^*} = \alpha\beta\boldsymbol{n_0} \tag{12}$$

Where $\alpha$ is same as section 3.1, $\beta$ is an adjustable factor, with the range of [0,1], and the optimal value is determined by experiments, which will be discussed extensively later.

The reconstructed decoder input is:

$$\boldsymbol{z_i^*} = \boldsymbol{z_i} + \boldsymbol{n_0^*} \tag{13}$$

## 4. Experiments

In this section, we have conducted several comparative experiments on MNIST, FashionMNIST [17] datasets, and FaceScrub[30] to demonstrate the effectiveness of CSAE, which contains more than 100,000 face-aligned images for 530 people, with 265 for men and women. We selected 50 men and 50 women, the number of samples in each class is close to 100 and all samples are flipped as training set. To facilitate comparison, we first designed the CSAE with convolution and CVAE with the same autoencoder structure, which is proposed in section 3, the dimension of the latent variable is 40. CSAE needs to consider classification performance, so that the SGD [27] optimization method is used for the optimizer, momentum is 0.9, and the weight decay is 0.0005. Thanks to batch normalization [26], we set the initial learning rate to 0.1, 120 training epochs, and reduce the learning rate by 10 times per 30 epochs. To match the input and output dimensions, we padded the training MNIST and FashionMNIST images to 32 × 32. The face image is resized to 128×128 after being cut by the ground truth bounding box.

In this way, different epochs are trained and the results of the models are compared. This section of the experiment is implemented under the Pytorch [12] framework. Implementations for CSAE can be found online at: https://github.com/anlongstory/CSAE

**4.1 Reconstruction performance**

We trained different epochs, CVAE converged quickly, we redesigned the loss function for CSAE, and increased the number of training epochs to 120. The images of MNIST test set is then input the model to reconstruction, as shown in Fig.4. There is a blurring phenomenon in CVAE. As the number of iterations increases, the CSAE reconstruction results become clearer. We add the Wavelets loss to make the results clearer. At the same time, adding noise is designed to make the network more robust to noise, which will blur the results to a certain extent. With the increase of $\beta$, the results of CSAE reconstruction gradually become blurred. But overall, CSAE reconstruction is still better than the CVAE which is under the same conditions.

In view of the above results, we did comparative experiments on Fashion-MNIST and FaceScrub too. As is shown in Fig.5 and Fig.6, For comparison, we directly give the results of the 120 training epochs, which β keeps the same value on FashionMNIST as MNIST. Because the latent variable dimension increases on FaceScrub, we reduce the selection range of β value. No matter reconstruction or generate random samples, CSAE are superior to CVAE obviously.

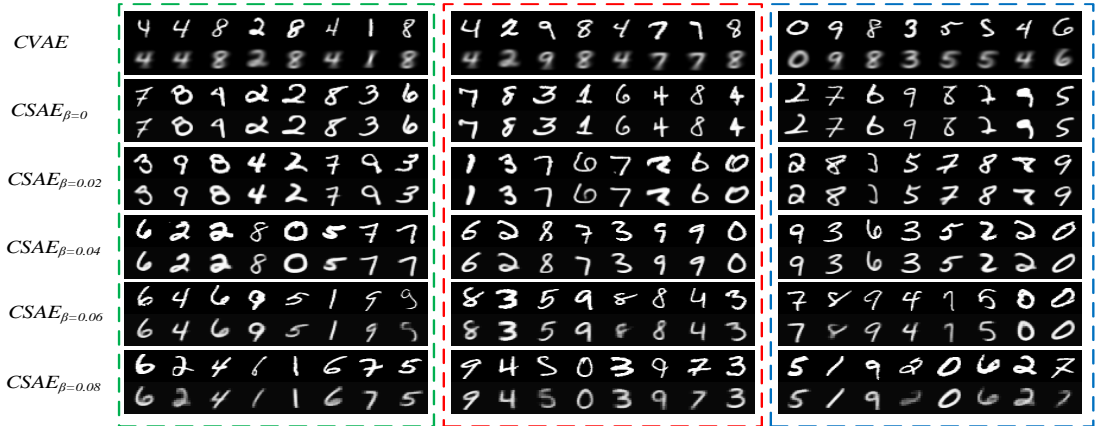

**Fig.4.** The reconstruction results of CVAE and CSAE with different $\beta$ on MNIST. Each set of results of the first row is ground-truth, the second row is model reconstruction results, where the green, red, and blue dotted frames represent the reconstructed results of the 40, 80, 120 training epochs, respectively.

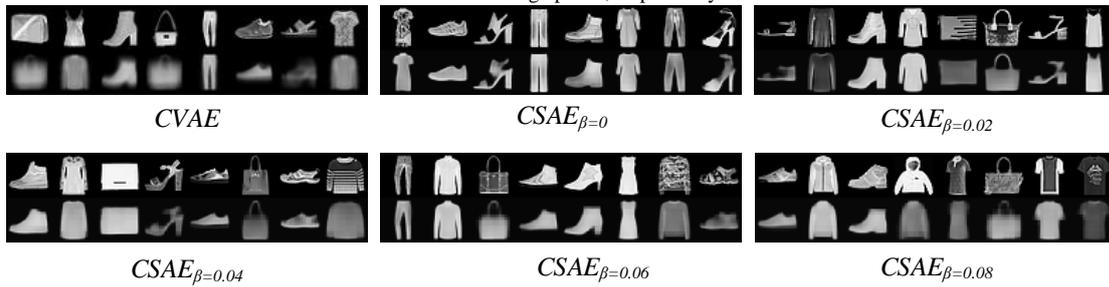

**Fig.5.** The reconstruction results of CVAE and CSAE with different $\beta$ on Fashion-MNIST, the first row is ground-truth, the second row is model reconstruction results.

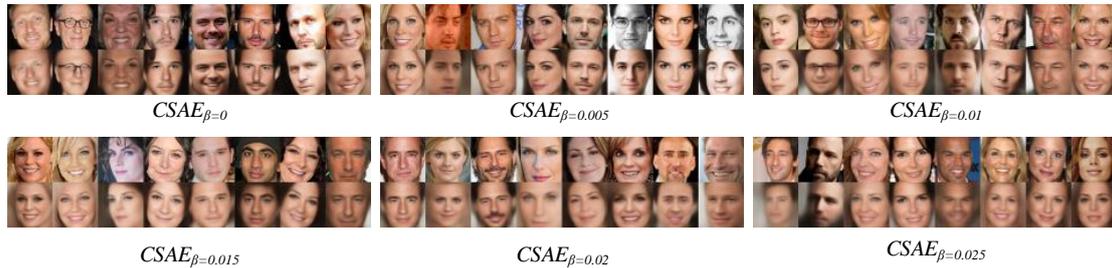

**Fig.6.** The reconstruction results of CSAE with different $\beta$ on FaceScrub. the first row is ground-truth, the second row is model reconstruction results.

### 4.2 Generating random samples

The mean and variance of latent variable are obtained through network learning in CVAE, and finally the random sample is generated from the data point subject to the standard normal distribution random value. Finally, through the reparameterization, it is input into the decoding network. Here we repeated for 10 times, each time 0-9 one-hot condition is added, 100 randomly generated samples were obtained. CSAE gets the 40-dimensional feature vectors of all training images in MNIST/FashionMNIST, then put the same class's feature vectors together, to calculate their mean and covariance matrix. According to the mean and covariance of each class, 10 random points are sampled to input into the decoding network. Finally, 100 random sample points were obtained.

With the gradual increase of the $\beta$, the generation of sample strokes has been gradually improved, indicating that the robustness of the model is improved, the sharpness of the generated numbers is becoming clearer, the quality of the generation is getting better. However, when $\beta$ is too large ($\beta = 0.08$), the model is more robust to the noise of the input latent vectors, but the decoded images also tend to be more standard form, resulting in a decrease in the diversity of samples. There is a contradiction between diversity and robustness, one side ascends, the other will decline.

Similarly, CSAE gets the 256-dimensional feature vectors of FaceScrub training images, 64 randomly generated samples were obtained. As shown in Fig.9, the upper left corner is the average face learned by the model. On the face data set, as β is increasing, the face is becoming more and more standardized, and the sharpness is gradually decreasing. In CVAE, when the training process is determined, it is difficult to control the choice between the diversity and robustness of the final generation of samples, while CSAE is different, where we can change the value of the $β$ coefficient in the Loss2 to tradeoff between diversity and robustness, so that training is more flexible.

As the number of iterations of the CSAE increases, the sharpness and completeness of the generated samples are improved, which also shows that Loss2 is constantly optimizing the model. The sharpness effect is better than the CVAE under the same condition.

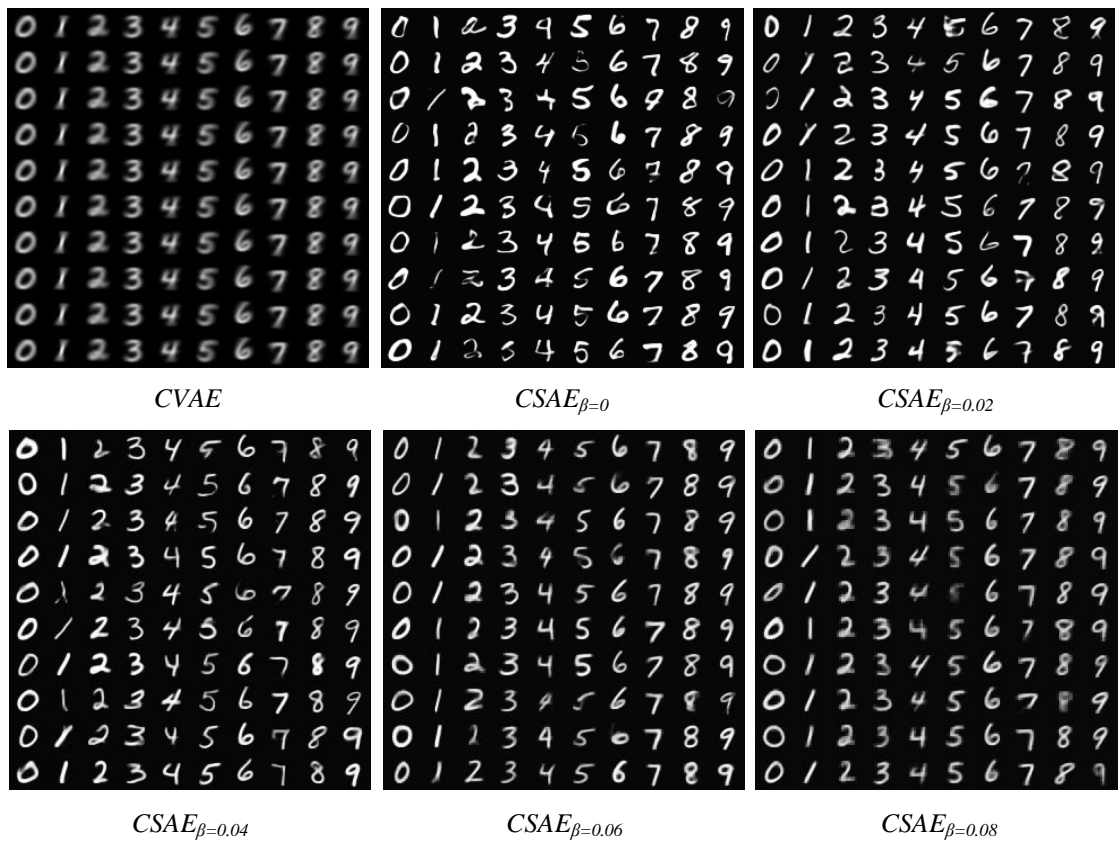

*CVAE*      $CSAE_{β=0}$      $CSAE_{β=0.02}$

$CSAE_{β=0.04}$      $CSAE_{β=0.06}$      $CSAE_{β=0.08}$

**Fig.7.** Randomly generated samples of CSAE with different $β$ on MNIST.

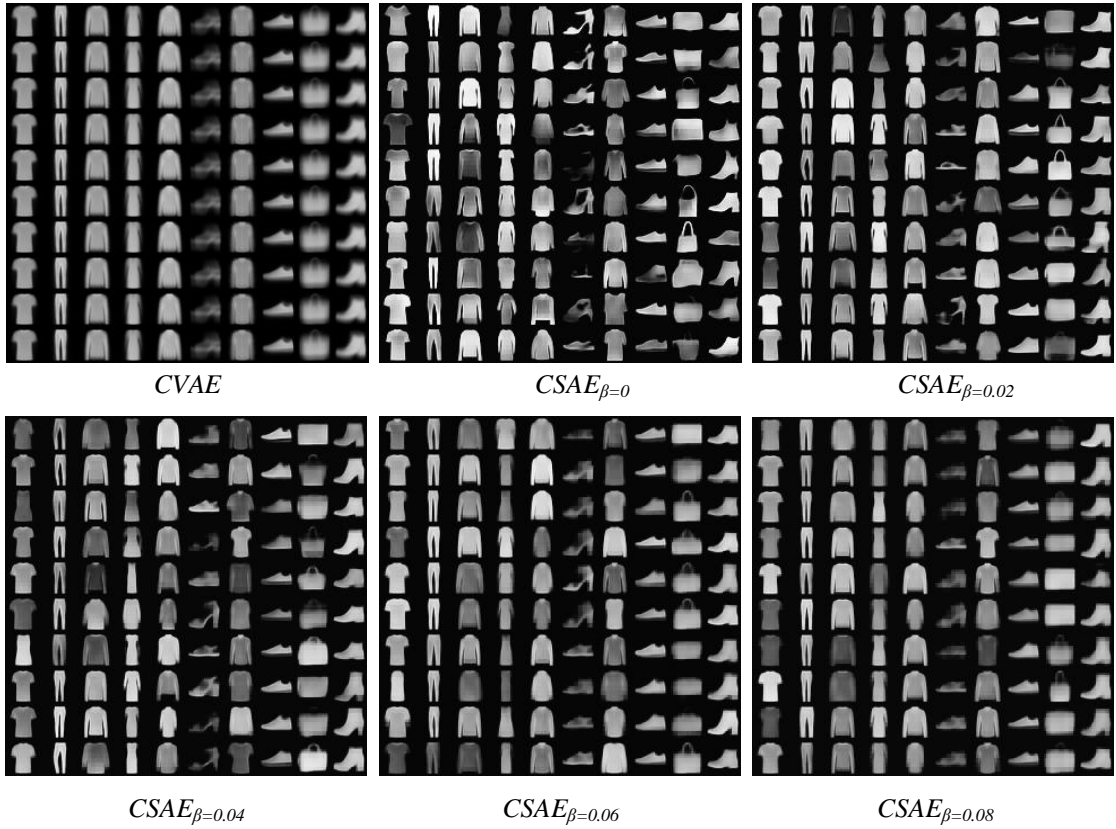
Fig.8. Randomly generated samples of CSAE with different $\beta$ on Fashion-MNIST

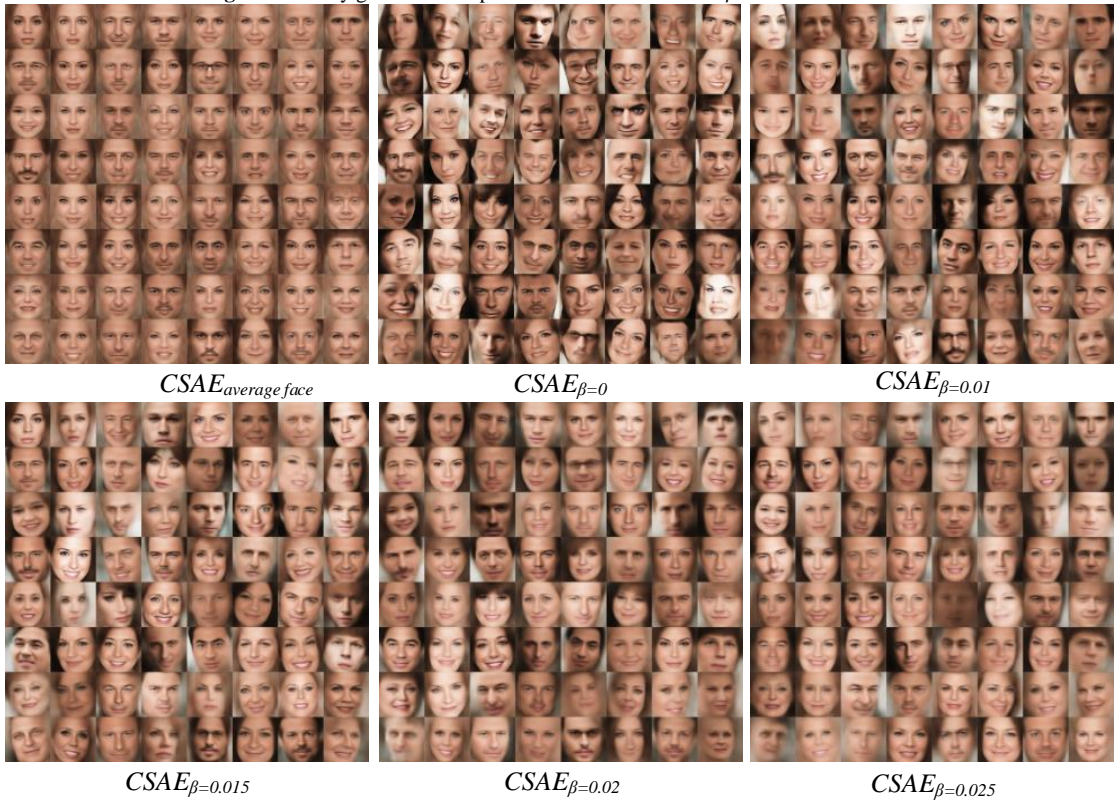
Fig.9. Randomly generated samples of CSAE with different $\beta$ on FaceScrub

### 4.3 Classification performance

CSAE's innovative linkage of latent variables directly to predefined class centroids, is used not only to generate samples, but also to classify patterns directly without adding a network structure. In Fig.10, the Euclidean distance between the cluster centers of latent variables are shown (because this distribution matrix is symmetrical, to make it easier to observe only the upper part is shown), the clustering center for numbers 9 and 7 should be relatively close [2,22], but due to the role of our predefined evenly-distributed class centroids, the minimum distance of the cluster center between each other is almost equal, which provides the best inter-class separation characteristics, improving the ability of pattern classification and extract features.

**Fig.10.** Euclidean Distance Matrix

Following is the analysis of network recognition performance on test dataset. CSAE uses the nearest class mean classifier for pattern classification, and predefined class center is mean value of each class. Table 2 shows the accuracy of classification on different datasets. Under the structure of the two fully connected layers, the recognition rate of nearly 98% can be achieved. The above results fully illustrate the feasibility of combining predefined class centers with latent variables to classify. To further compare the classification performance of CSAE, we design a comparative experiment in which the $CSAE_{encoder}$ only contains the encoder part of CSAE with nearest class mean classifier. $Basenet$ is the convolution neural network which is consistent with the structure of the CSAE encoder in the convolution part, and only the last fully connected layer is replaced with the $4 \times 4$ avepooling layer and Softmax layer, the original label and cross entropy loss function is used for training.

**Table 3** The classification results on MNIST/FashionMNIST

| Model | Recognition rate | |
| --- | --- | --- |
|  | MNIST | FashionMNIST |
| $CSAE_{\beta=0}$ | 99.48% | 92.70% |
| $CSAE_{\beta=0.02}$ | **99.52%** | 92.36% |
| $CSAE_{\beta=0.04}$ | 99.50% | 92.45% |
| $CSAE_{\beta=0.06}$ | 99.50% | **92.89%** |
| $CSAE_{\beta=0.08}$ | 99.47% | 92.48% |
| $CSAE_{encoder}$ | 99.45% | 92.60% |
| Basenet | 99.20% | 92.45% |

As you can see from Table 3 and Table 4, the networks of CSAE structure, whether CSAE with different $\beta$ or $CSAE_{encoder}$ can be better than convolution neural networks that use cross-entropy training. Compared with $CSAE_{encoder}$, the CSAE classification performance is also improved, which shows that the decoder actually improves the classification performance. In addition, due to the use of convolution neural network, the number of parameters of the network is obviously reduced, and the performance is also obviously improved, which shows that the network structure has a great impact on the performance of autoencoder.

**Table 4** The classification results on FaceScrub

| Model | Recognition rate |
| --- | --- |
|  | FaceScrub |
| $CSAE_{\beta=0}$ | 92.10% |
| $CSAE_{\beta=0.005}$ | 92.69% |
| $CSAE_{\beta=0.01}$ | 92.88% |
| $CSAE_{\beta=0.015}$ | 92.80% |
| $CSAE_{\beta=0.02}$ | **92.84%** |
| $CSAE_{encoder}$ | 92.08% |
| Basenet | 90.18% |

### 4.4 Generalization performance

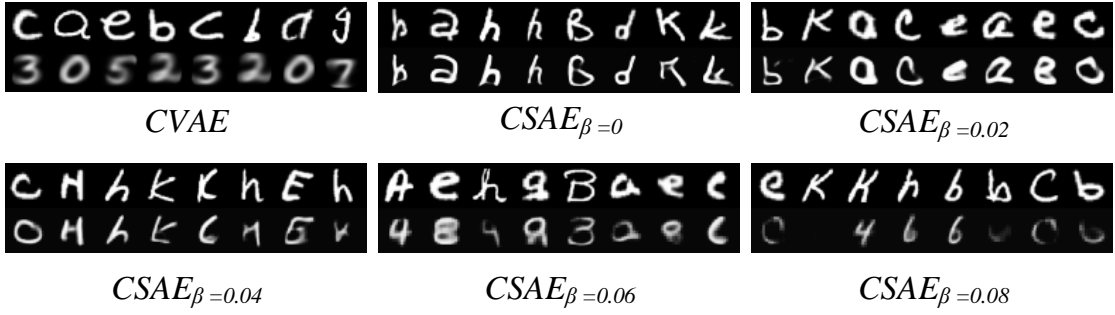

*CVAE*    *CSAE$_{\beta=0}$*    *CSAE$_{\beta=0.02}$*

*CSAE$_{\beta=0.04}$*    *CSAE$_{\beta=0.06}$*    *CSAE$_{\beta=0.08}$*

**Fig.11.** The reconstruction results of CVAE and CSAE with different $\beta$ on EMNIST, the first row is ground-truth, the second row is model reconstruction results.

To verify the generalization ability of the model, we test the extrapolation ability of the model in experiments, while its interpolation ability is embodied in the generated sample quality and network classification performance mentioned earlier. For the model trained on the MNIST dataset, we randomly picked 10 classes from the letter subset of the EMNIST [23] dataset, the model has never seen a sample of characters, the samples are input directly into the model for reconstruction. CVAE reconstructed letters belong only to MNIST classes, it cannot rebuild a class that have not seen at all. In CSAE, we test the $\beta$ which takes different values including $\beta = 0$, i.e. no random noise adding. When the random noise is not added, the characters can be reconstructed, and with the increase of $\beta$, the reconstruction of characters becomes more and more blurred, and when the final $\beta = 0.08$, It is also difficult for CSAE to reconstruct letters that have not been seen. Here, because CSAE changes the combination of the loss function, depending on the different usage scenario (generate a more robust model or obtain a more generalized model), we can artificially adjust the generalization ability of the model by changing $\beta$. To sum up, through $\beta$, we can adjust the performance of CSAE in interpolation and extrapolation, more flexible than CVAE.

### 4.5 Loss function

During training, different losses have their importance. In order to further illustrate the effectiveness of the Wavelets loss function, we chose the CSAE model with $\beta = 0.005$ and trained on the faceScrub dataset, using MSE loss and Wavelets loss, respectively. After training, we made the following comparison, and can see that whether the reconstruction results or generation samples, wavelets loss is clearer than MSE loss. In terms of accuracy, the accuracy of MSE is 92.05%, while the accuracy of wavelets loss is 92.69%. Each indicator illustrates the effectiveness of wavelets loss.

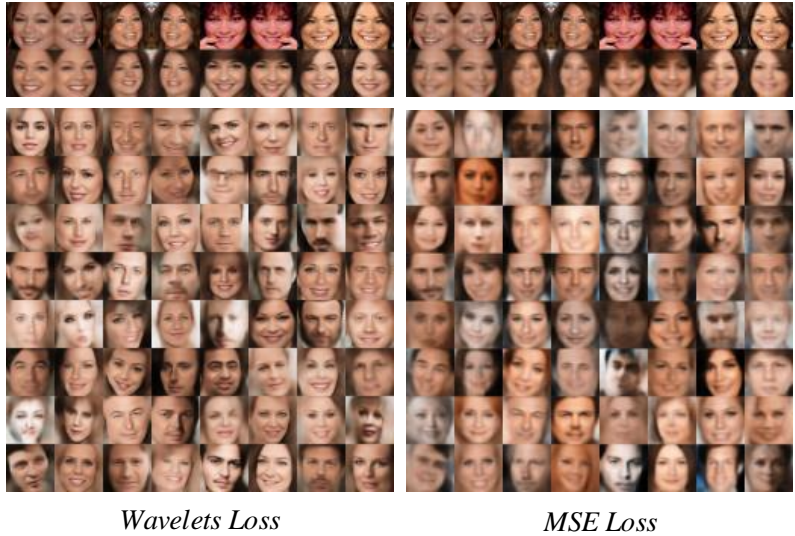

*Wavelets Loss*    *MSE Loss*

**Fig.12.** The results of different loss functions

## 5. Conclusions

This paper mainly introduces a new autoencoder structure named CSAE, by setting a predefined evenly-distributed class centroids (PEDCC) so that the autoencoder has the classification function at the same time, that is, the intermediate latent variable obtained by the encoder is used not only for decoding, but also directly for classification. PEDCC guarantee the largest distance between different classes center, and make it easier to classification and generate samples. Theoretically, it can generate any number of samples by sampling within different class of distribution. In our work, a new loss function is constructed, which is more flexible in parameter selection. We can artificially change the coefficients in training to control the final reconstruction results and the quality of the samples generated, and validates this statement by means of visual display and numerical analysis. The future works are to explore CSAE in more complex datasets, such as higher quality faces and natural images. In addition, due to the and broader tasks such as incremental learning, semantic segmentation, style transfer and so on.

**Reference**

[1] Kingma D P, Welling M. Auto-encoding variational bayes[J]. arXiv preprint arXiv:1312.6114, 2013.

[2] Doersch C. Tutorial on variational autoencoders[J]. arXiv preprint arXiv:1606.05908, 2016.

[3] Kingma D P, Mohamed S, Rezende D J, et al. Semi-supervised learning with deep generative models[C]//Advances in neural information processing systems. 2014: 3581-3589.

[4] Liu W, Wen Y, Yu Z, et al. Large-Margin Softmax Loss for Convolutional Neural Networks[C]//ICML. 2016: 507-516.

[5] Wang F, Cheng J, Liu W, et al. Additive margin softmax for face verification[J]. IEEE Signal Processing Letters, 2018, 25(7): 926-930.

[6] Howard A G, Zhu M, Chen B, et al. Mobilenets: Efficient convolutional neural networks for mobile vision applications[J]. arXiv preprint arXiv:1704.04861, 2017.

[7] Zhang X, Zhou X, Lin M, et al. Shufflenet: An extremely efficient convolutional neural network for mobile devices[C]//Proceedings of the IEEE Conference on Computer Vision and Pattern Recognition. 2018: 6848-6856.

[8] He K, Zhang X, Ren S, et al. Deep residual learning for image recognition[C]//Proceedings of the IEEE conference on computer vision and pattern recognition. 2016: 770-778.

[9] Simonyan K, Zisserman A. Very deep convolutional networks for large-scale image recognition[J]. arXiv preprint arXiv:1409.1556, 2014.

[10] Hu J, Shen L, Sun G. Squeeze-and-excitation networks[C]//Proceedings of the IEEE Conference on Computer Vision and Pattern Recognition. 2018: 7132-7141.

[11] Deng L. The MNIST Database of Handwritten Digit Images for Machine Learning Research [Best of the Web][J]. IEEE Signal Processing Magazine, 2012, 29(6):141-142.

[12] Paszke A, Chintala S, Collobert R, et al. Pytorch: Tensors and dynamic neural networks in python with strong gpu acceleration, may 2017[J].

[13] Olshausen B A, Field D J. Emergence of simple-cell receptive field properties by learning a sparse code for natural images[J]. Nature, 1996, 381(6583): 607.

[14] Lee H, Battle A, Raina R, et al. Efficient sparse coding algorithms[C]//Advances in neural information processing systems. 2007: 801-808.

[15] Vincent P, Larochelle H, Bengio Y, et al. Extracting and composing robust features with denoising autoencoders[C]//Proceedings of the 25th international conference on Machine learning. ACM, 2008: 1096-1103.

[16] Bengio Y, Yao L, Alain G, et al. Generalized denoising auto-encoders as generative models[C]//Advances in Neural Information Processing Systems. 2013: 899-907.

[17] Xiao H, Rasul K, Vollgraf R. Fashion-mnist: a novel image dataset for benchmarking machine learning algorithms[J]. arXiv preprint arXiv:1708.07747, 2017.

[18] Rezende D J, Mohamed S, Wierstra D. Stochastic backpropagation and approximate inference in deep generative models[J]. arXiv preprint arXiv:1401.4082, 2014.

[19] Sohn K, Lee H, Yan X. Learning structured output representation using deep conditional generative models[C]//Advances in Neural Information Processing Systems. 2015: 3483-3491.

[20] Walker J, Doersch C, Gupta A, et al. An uncertain future: Forecasting from static images using variational autoencoders[C]//European Conference on Computer Vision. Springer, Cham, 2016: 835-851..

[21] Lavda F, Ramapuram J, Gregorova M, et al. Continual Classification Learning Using Generative Models[J]. arXiv preprint arXiv:1810.10612, 2018.

[22] Kang W Y, Zhang B T. "Continual learning with Generative replay via discriminative variationalautoencoder"[Online].Available:https://marcpickett.com/cl2018/CL-2018_paper_15.pdf

[23] Cohen G, Afshar S, Tapson J, et al. EMNIST: an extension of MNIST to handwritten letters[J]. arXiv preprint arXiv:1702.05373, 2017.

[24] Pandey G, Dukkipati A. Variational methods for conditional multimodal learning: Generating human faces from attributes[R]. Technical Report, 2016.

[25] McCloskey M, Cohen N J. Catastrophic interference in connectionist networks: The sequential learning problem[M]//Psychology of learning and motivation. Academic Press, 1989, 24: 109-165.

[26] Ioffe S, Szegedy C. Batch normalization: Accelerating deep network training by reducing internal covariate shift[J]. arXiv preprint arXiv:1502.03167, 2015.

[27] Kingma D P, Ba J. Adam: A method for stochastic optimization[J]. arXiv preprint arXiv:1412.6980, 2014.

[28] Optimizing the Latent Space of Generative Networks 2017(GLO)

[29] Non-Adversarial Image Synthesis with Generative Latent Nearest Neighbors

[30] H. Ng and S. Winkler, "A data-driven approach to cleaning large face datasets," in 2014 IEEE International Conference on Image Processing (ICIP), 2014, pp. 343–347.